\documentclass[twoside]{article}

\usepackage[accepted]{aistats2025}

\usepackage{graphicx} 
\usepackage{mathtools}
\usepackage{hyperref}
\usepackage{amssymb}
\usepackage{amsthm}
\usepackage{xcolor}
\newtheorem{theorem}{Theorem}[section]

\usepackage{booktabs}
\begin{document}
\twocolumn[

\aistatstitle{Making Sense Of Distributed Representations With Activation Spectroscopy}

\aistatsauthor{ Kyle Reing \And Greg Ver Steeg \And  Aram Galstyan }

\aistatsaddress{ University of Southern California \And  University of California, Riverside \And University of Southern California } ]

\begin{abstract}
    In the study of neural network interpretability, there is growing evidence to suggest that relevant features are encoded across many neurons in a distributed fashion. Making sense of these distributed representations without knowledge of the network's encoding strategy is a combinatorial task that is not guaranteed to be tractable. This work explores one feasible path to both detecting and tracing the joint influence of neurons in a distributed representation. We term this approach \textit{Activation Spectroscopy} (ActSpec), owing to its analysis of the pseudo-Boolean Fourier spectrum defined over the activation patterns of a network layer. The sub-network defined between a given layer and an output logit is cast as a special class of pseudo-Boolean function. The contributions of each subset of neurons in the specified layer can be quantified through the function's Fourier coefficients. We propose a combinatorial optimization procedure to search for Fourier coefficients that are simultaneously high-valued, and non-redundant. This procedure can be viewed as an extension of the Goldreich-Levin algorithm which incorporates additional problem-specific constraints. The resulting coefficients specify a collection of subsets, which are used to test the degree to which a representation is distributed. We verify our approach in a number of synthetic settings and compare against existing interpretability benchmarks. We conclude with a number of experimental evaluations on an MNIST classifier, and a transformer-based network for sentiment analysis.
\end{abstract}

\section{INTRODUCTION}

Given access to neural recordings in any layer of a neural network, is it possible to discern what input features are represented, and how these features are encoded? These questions are at the heart of the ongoing crisis in neural network interpretability, and it's unclear whether an affirmative answer exists, even under highly simplifying assumptions. If features are encoded independently using individual neurons, then understanding the representation becomes a simple matter of deciphering the role of each neuron. From the perspective of interpretability, this would be the best case scenario, and a plethora of work exists which considers this assumption as axiomatic.  However, there's reason to believe that such a straight-forward encoding strategy is not utilized by networks in general. For starters, individual neurons have frequently been observed to respond to multiple disparate features, such as cats and the front of cars \cite{poly}. While it's possible that these examples share a subtle visual feature, a more likely explanation is that multiple features utilize an individual neuron in their encoding. Such neurons have been referred to as polysemantic \cite{poly2}, and their existence provides evidence for a distributed encoding strategy. In a distributed encoding, features are encoded using the joint state of some collection of neurons, whereby each neuron specifies a coordinate in a larger feature space rather than (or in addition to) being individually meaningful. For a simple example of a distributed encoding, one could look at any letter of this text, which is represented jointly on the underlying hardware using the 8-bit ASCII encoding strategy.\\

If the encoding strategy that neural networks stumble onto is distributed, what sort of challenges does that present to interpretability? If any joint collection of neurons could potentially represent a feature, the problem of interpretability becomes exponentially harder. Without prior knowledge, each subset of $n$ variables becomes a potential candidate whose role - if any - must be determined. All this is not even considering the possibility that many features may be encoded in different regions of the continuous space spanned by some subset of neurons. Needless to say, the general case becomes woefully intractable. In light of this, is there any hope for interpretability if the underlying representation is distributed? Is it even possible to determine whether a representation is distributed or not? In this work we bring attention to these questions, and provide one possible way forward. We do this by casting the search for relevant subsets as a combinatorial optimization problem. By representing a neural network with a very specific type of pseudo-Boolean function, we are able to perform this search very efficiently. After an introduction to relevant neighboring literature in Section 1, we spend most of Section 2 and 3 setting up how a neural network can be meaningfully represented as a pseudo-Boolean function. Once the connection is established, we commit the rest of the paper to performing experiments which showcase how this combinatorial search procedure can be leveraged towards problems in interpretability. 

\section{RELATED WORK}
Work in interpretability over the last decade has been in constant battle between what is correct and what is efficient. The first methods to be proposed, such as saliency maps, used readily available byproducts of the network, but produced results of questionable utility \cite{unreli}. Similarly with language models, components like attention which were once used as unquestionable windows into the model are now being brought into question \cite{attn1}\cite{attn2}. Continuing with this trend is the recent push towards mechanistic interpretability.\cite{mono}\cite{mono_scale} Such work aims to look beyond simple metrics like which neurons have the largest activation, and instead into the underlying mechanisms that make the network function. This work sits squarely under the umbrella of mechanistic interpretability since it aims to look at the network across the entire dataset and tease out which subspaces of neurons are working together to yield the output. Where this work stands apart is its use of higher-order analysis through the Fourier spectrum. Higher-order importance has been studied \cite{sti}, but to the best of our knowledge has not overcome the combinatorial hurdle associated with measuring over subsets. Spectroscopy has been used in interpretability, such as in logit spectroscopy \cite{logit} to further study the logit lens used for LLMs. However, this often refers to the eigenspectrum of a matrix instead of the Fourier spectrum of a pseudo-Boolean function.

\begin{figure}
    \centering
    \includegraphics[scale=1.1]{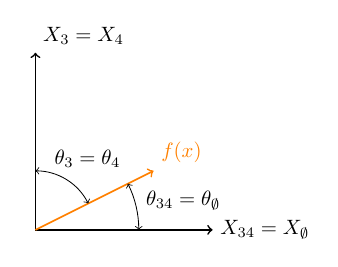}
    \caption{Showcasing redundant variables, where a large inner product between variables leads to identical Fourier coefficients}
    \label{fig:red}
\end{figure}

Within the context of machine learning this is not the first time a variant of pseudo-Boolean Fourier analysis has been used. The classic algorithm of (Mansour, \cite{km}) is utilized to this end to learn decision trees. However, our approach does not do any learning, and instead analyses a pre-existing network that has already been trained without such limiting constraints. Additionally, we are not directly applying the classical Goldreich-Levin algorithm \cite{gl} in our analysis. Instead of simply looking for the highest Fourier coefficients, we are looking for high Fourier coefficients that also satisfy additional conditions (discussed in Section 4).

\section{PREREQUISITES}

\subsection{Pseudo-Boolean Functions}
A pseudo-Boolean function is a function of the form $f: \mathbb{B}^n \rightarrow \mathbb{R}$, where $\mathbb{B}$ is a Boolean domain. In this work we consider $\mathbb{B} = \{-1, 1\}$, since under this assumption the $2^n$ parity functions $x^S = \prod\limits_{i \in S} x_i$ for all $S \subseteq [n]$ form an orthonormal basis \cite{odonnell}. This implies that each pseudo-Boolean function has a unique representation, which can be stated with the following Fourier Expansion Theorem:

\begin{table*}
    \centering
    \begin{tabular}{c|c|c|c|c|c}
        \textbf{Approach} & \textbf{Hard-coded} & \textbf{Learned} & \textbf{Constant} & \textbf{Noise} $\mathbf{(1000)}$ & \textbf{Noise} $\mathbf{(50)}$\\
        \midrule
         \textbf{Integrated Gradients} & $0.1562$ & $0.0507$ & $0.3885$& $0.3766$ & $0.4674$ \\
         \midrule
         \textbf{Gradient Shap} & $0.1562$ & $0.0864$ & $0.3354$& $0.3756$ & $0.3136$\\
         \midrule
         \textbf{InputXGradient} & $0.1562$ & $0.1051$ & $0.3711$& $0.3522$  &  $0.3378$ \\
         \midrule
         \textbf{Saliency} & $\mathbf{0.0794}$ & $0.0403$ & $\mathbf{0.1656}$& $\mathbf{0.2460}$ &  $\mathbf{0.2498}$\\
         \midrule
         \textbf{Guided Backprop} & $0.1850$ & $0.0719$ & $0.3905$& $0.6959$ & $0.6930$\\
         \midrule
         \textbf{Feature Ablation} & $0.1562$ & $0.0718$ & $0.4212$& $0.4410$ & $0.4852$ \\
         \midrule
         \textbf{Shapley Values} & $0.1562$ & $0.0517$ & $0.4751$& $-$ & $-$\\
         \midrule
         \textbf{Shapley Sampling} & $0.1562$ & $\mathbf{0.0312}$ & $0.4688$& $-$ &  $-$\\
         \midrule
         \textbf{ActSpec} & $\mathbf{0.0}$ & $\mathbf{0.0}$ & $\mathbf{0.0}$ & $\mathbf{0.0424}$ & $\mathbf{0.1875}$ \\
    \end{tabular}
    \caption{Total variation distance from the ground truth importance for a number of synthetic settings and approaches.}
    \label{tab:synth}
\end{table*}

\begin{theorem}[Fourier Expansion Theorem]
    Every function \(f:\{-1,1\}^n \rightarrow \mathbb{R} \) can be uniquely expressed as a multilinear polynomial,
    \begin{equation}
        f(x) = \sum\limits_{S \subseteq [n]} \hat{f}(S)x^S.
    \end{equation}
    This expression is called the Fourier expansion of \(f\), and the real number \( \hat{f}(S)\) is called the Fourier coefficient of \(f\) on \(S\).
    \label{thm:31}
\end{theorem}

These Fourier coefficients are computed using the following inner product, which is often normalized:

\begin{equation*}
    \hat{f}(S) = 2^{-n} \langle f, x^S \rangle = 2^{-n} \sum_{\mathclap{x \in \{-1,1\}^n}} \big(f(x)\cdot \prod_{i \in S}x_i \big)
\end{equation*}

Geometrically, these Fourier coefficients can be thought of as a function of the angle between two $2^n$ dimensional vectors. One vector is the function output $f(x)$, and the other is the parity $x^S$, with each coordinate of the vector corresponding to the evaluation of a Boolean input in $\{-1,1\}^n$. More specifically, the following equation holds:
\begin{equation*}
    \langle f, x^S \rangle = cos(\theta_S)|x^S||f(X)|
\end{equation*} where the length $|A| = \sqrt{\sum A^2}$. If the function output is also Boolean, then each Fourier coefficient is just the cosine distance between the parity of a subset and function output. In the general case, a pseudo-Boolean Fourier coefficient can be thought of as a scaled cosine distance, based on the length $|f(x)|$.

\subsection{Goldreich-Levin Algorithm}
Computing the exact Fourier decomposition for an arbitrary pseudo-Boolean function is not generally tractable. However, the behavior of a function is largely determined by it's high-valued Fourier coefficients \cite{odonnell}. This means if the Fourier spectrum of a function is sufficiently concentrated, it is enough to only estimate the high-valued coefficients. The Goldreich-Levin algorithm \cite{gl} is a well established approach for finding and subsequently estimating these coefficients. The algorithm accomplishes this by taking advantage of two facts. The first is that the sum over all squared Fourier coefficients is equal to some constant $\mathcal{C}$. This defines a density function over all subsets $S$ such that $\hat{f}(S)^2$ measures the contribution of $S$ to the total weight $\mathcal{C}$. This allows for a threshold $\gamma^2$ to be defined such that any subset with a contribution exceeding this threshold is accepted by the algorithm. The second fact is that sums over collections of squared Fourier coefficients can be estimated accurately and efficiently:\\

\begin{theorem}[Goldreich-Levin Theorem \cite{odonnell}]
For arbitrary subsets $I \subseteq [n]$, $J = [n] / I$, $S \subseteq I$, and $T \subseteq J$, the sum over squared coefficients $\sum_{T} \hat{f}(S\cup T)^2$ can be estimated to error $\pm \eta$ with probability $1 - \delta$ using only $O(\log(\frac{1}{\delta} / \eta^2))$ samples.
\end{theorem}

Together, these two facts contribute to a branch-and-bound style combinatorial optimization algorithm, where candidate subsets are discarded if their estimated cumulative weight does not exceed a threshold.

\begin{figure*}
    \centering
    \includegraphics[scale=0.3]{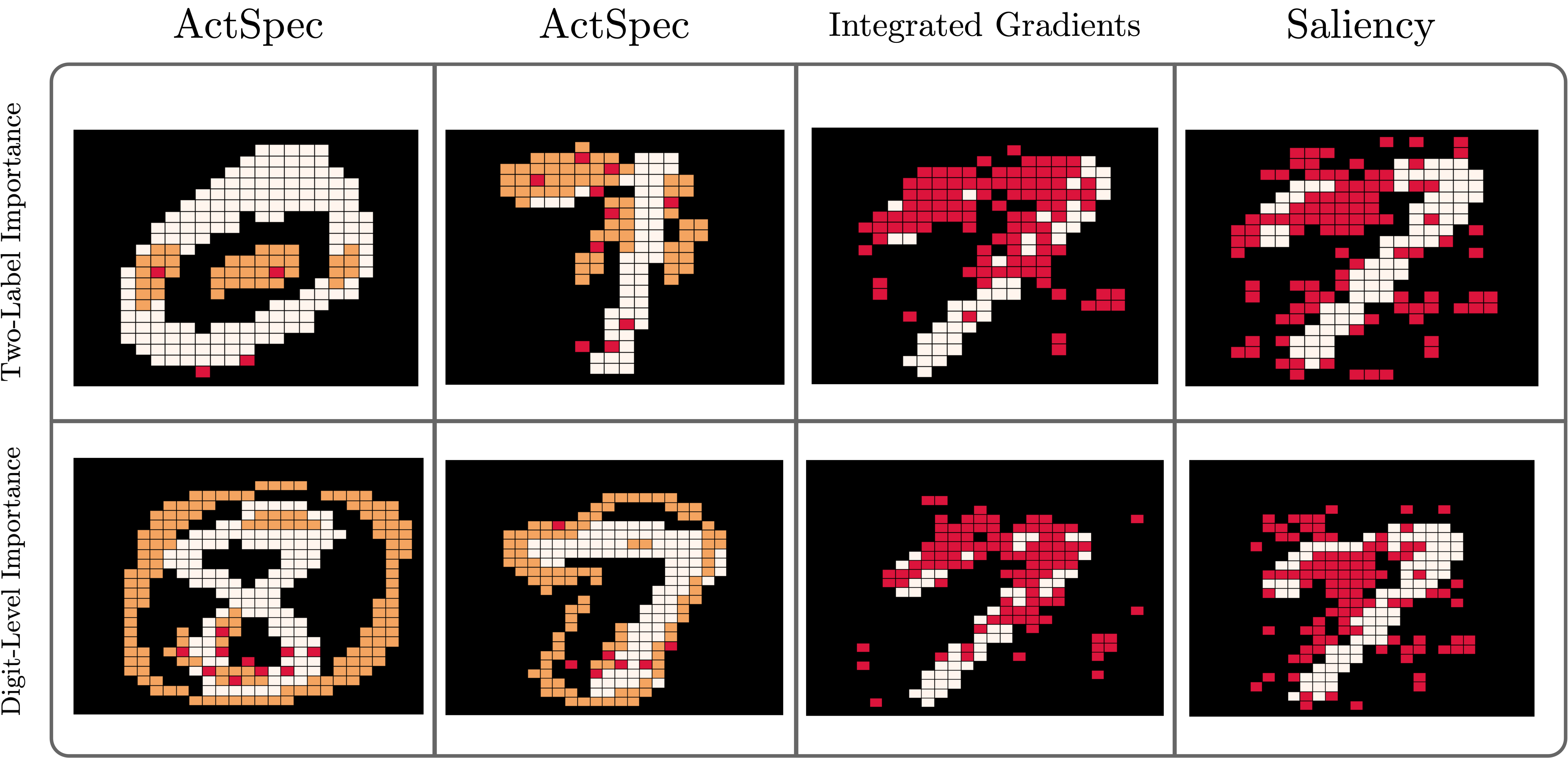}
    \caption{Top Row: Experiments on samples from two classes. Bottom Row: Experiments on all of MNIST. Left: ActSpec (red pixels) alongside redundancy (orange pixels). Right: Attribution methods for the same task.}
    \label{fig:mnist_input}
\end{figure*}

\section{METHODS}
Two major problems exist which prevent a direct application of the Goldreich-Levin algorithm to neural network interpretability. Addressing these concerns is the core theoretical contribution underlying Activation Spectroscopy. The first issue is one of Out-of-distribution sampling, and the second is Fourier Coefficient redundancy.

\subsection{Out-of-distribution Sampling}
Pseudo-Boolean functions are defined over the entire Boolean hypercube $\{ -1, 1\}^n$, and existing methods for analysis of these functions rely on taking uniformly random samples over this space. In the context of neural networks, while inputting a uniformly random sample will produce an output, such an evaluation is practically useless in informing about the normal operating conditions under which the network was trained. To better reflect the underlying behavior, the sample space can be divided into In-distribution and Out-of-distribution collections. The relevance of pseudo-Boolean function analysis thus depends on its ability to meaningfully operate in the In-distribution setting. We show that for a special class of pseudo-Boolean functions, which we term pseudo-Boolean projections, uniform sampling over an In-distribution subspace is equivalent to uniform sampling over the entire space up to a constant factor. This implies that all existing guarantees still hold in the In-distribution setting if the underlying object is assumed to be a projection.

We sketch out the high-level formulation here, with a more in-depth discussion relegated to the Appendix. Informally, we consider the set of pseudo-Boolean functions that output $0$ for any sample that is Out-of-distribution. For a pseudo-Boolean function $f(X)$, the $2^n$ outputs define a vector space $V \in \mathbb{R}^{2^n}$. Define the projection $P:V \rightarrow V$ such that $V_m = 0$ for sample $m$ if $m \not\in \mathbb{D}$, otherwise $f(X)$ is unchanged. Such an operation is an orthogonal projection from $\mathbb{R}^{2^n}$ onto $\mathbb{R}^D$, where $D$ is the number of In-distribution samples. The resulting function $g(X) = P\cdot f(X)$ is still a pseudo-Boolean function, and thus by Theorem \ref{thm:31} has a unique multilinear expansion. In general, defining $2^n$ Fourier coefficients from $D$ samples is underdetermined and has infinitely many solutions. However, the projection is the only such function to satisfy two additional conditions. The first is that the Fourier coefficients on the subspace $\mathbb{R}^D$ match those of the projection on the full space $\mathbb{R}^{2^n}$:
\begin{equation*}
    \begin{split}
        \hat{g}(S) &= \hat{f}(S) - 2^{-n} \sum\limits_{x \notin \mathbb{D}} f(x)\cdot \prod_{i \in S}x_i\\
        &= 2^{-n} \sum\limits_{x \in \mathbb{D}} f(x)\cdot \prod_{i \in S}x_i
    \end{split}
\end{equation*}


The second is that Out-of-distribution samples do not contribute in any way to the Fourier coefficients of the projection, which cannot be said for non-zero values. As such it is similar to a maximum entropy condition, whereby uncertainty is maximized outside of the satisfied constraints. Sample guarantees for Goldreich-Levin are obtained via the Hoeffding inequality and uniform sample draws. However, uniformly sampling the non-zero values of an expectation is proportional for some normalization constant to uniform sampling over the full space. Thus the sample guarantees for Goldreich-Levin also apply to pseudo-Boolean projections.
\subsection{Fourier Coefficient Redundancy}
Given a pseudo-Boolean projection $f(X)$, a new set of challenges arises with respect to the interpretation of Fourier Coefficients. To illustrate, consider the following example in Table\ref{tab:my_label}:

\begin{table}[h]
    \centering
    \begin{tabular}{c|c|c|c|c|c}
        $X_1$ & $X_2$ & $X_3$ & $X_4$ & $X_5$&$f(X)$\\
        \midrule
         1& 1 &  1& 1&1 & 1\\
         1& -1 &  -1& -1&1 & -1\\
         -1&  1&  -1& -1&1& -1\\
         -1&  -1&  1& 1&1 & 1\\
         \midrule
         - & -& - & - &-& 0
    \end{tabular}
    \label{tab:my_label}
\end{table}
In this case we have $4$ In-distribution samples, and the remaining $28$ are Out-of-distribution. In terms of the Fourier coefficients for this function, one might expect high values for $\hat{f}(X_1X_2)$, $\hat{f}(X_3)$, and $\hat{f}(X_4)$. Instead, a number of large Fourier coefficients exist for large collections of variables, such as $\hat{f}(X_1X_2X_3X_4X_5)$, despite the function being seemingly predictable from a single variable. Returning to the definition of the Fourier coefficients in Section 3.1, it's easy to see each coefficient is a scaled cosine distance between a parity of variables and the output. Since only the In-distribution samples contribute to the coefficients, multiple variables can become linearly dependent with respect to their non-zero contributions. When two vectors are the same in their non-zero samples, such as $X_3$ and $X_4$, their cosine distances are the same, and are thus interchangeable in any subset. Additionally, the parity of two identical variables is constant.

As illustrated in Figure \ref{fig:red} this issue arises when individual variables or subsets are redundant with each other, which happens when their inner product is non-zero when conditioning on In-distribution samples. The reason this is an issue is that a redundancy may lead one to believe that a complex relationship is present, when in reality there is none. In the above example, since $X_3$ is the same In-distribution vector as $X_1\cdot X_2 \cdot X_4 \cdot X_5$, a minimal description is preferable. Thus the goal of ActSpec is to estimate high-valued Fourier coefficients which are also minimal in the sense that the vectors cannot be redundantly explained with simpler collections of variables. It should be noted however that keeping track of the redundant relationships present is very useful, and will play a role in later experiments.


In order to ensure that a variables contribution to some Fourier coefficient is not redundant, we modify the classical Goldreich-Levin algorithm in the following way. Under normal circumstances, a variable is added for consideration in a subset, and the sum over Fourier coefficients containing that variable is estimated. If that estimate exceeds the threshold $\tau$, the variable is accepted and the algorithm proceeds. ActSpec adds an additional filtering step, where we check to see if the inner product between the new variable and all subsets of existing variables in the set is small. If the inner product is not small, this implies that the new variable was about to be accepted simply because it was redundant. Although it may seem that estimating the inner product between an exponential number of vectors is infeasible, we're able to use the same trick as the Goldreich-Levin algorithm, but in reverse. Goldreich-Levin is able to estimate the sum over an exponential number of inner products, and check to see if this sum is above a threshold. Since all of the variables under consideration in our redundancy test are Boolean, we can do the same thing, except ensuring that the sum is below a new threshold $\gamma$. In total the run time of the algorithm is only a constant multiple of the original, making it $O(2nS \log\frac{1}{\delta})$, where $S$ is the number of subsets above $\tau$. In practice the value of $S$ is strictly smaller for ActSpec since it filters additional variables.

With the issues of Out-of-distribution sampling and Redundant Fourier Coefficients addressed, we can now proceed to use ActSpec for the problems of neural network interpretability. For a given neural network layer, we consider the intermediate activation patterns as the input to some pseudo-Boolean projection onto In-distribution samples. We then run ActSpec on this function, which searches for high-valued and non-redundant Fourier coefficients that contribute the most to the functions output. While this approach relies on looking at the binarized state of some latent space, which has the potential to be very lossy (such as in the case of a dense continuous encoding), for every experiment we ran, we found that all joint activation patterns corresponded to a unique continuous representation vector. While this supports the one-to-one mapping in our experiments, we are unsure how generally applicable this is across all networks.

\begin{figure*}
    \centering
    \includegraphics[scale=0.3]{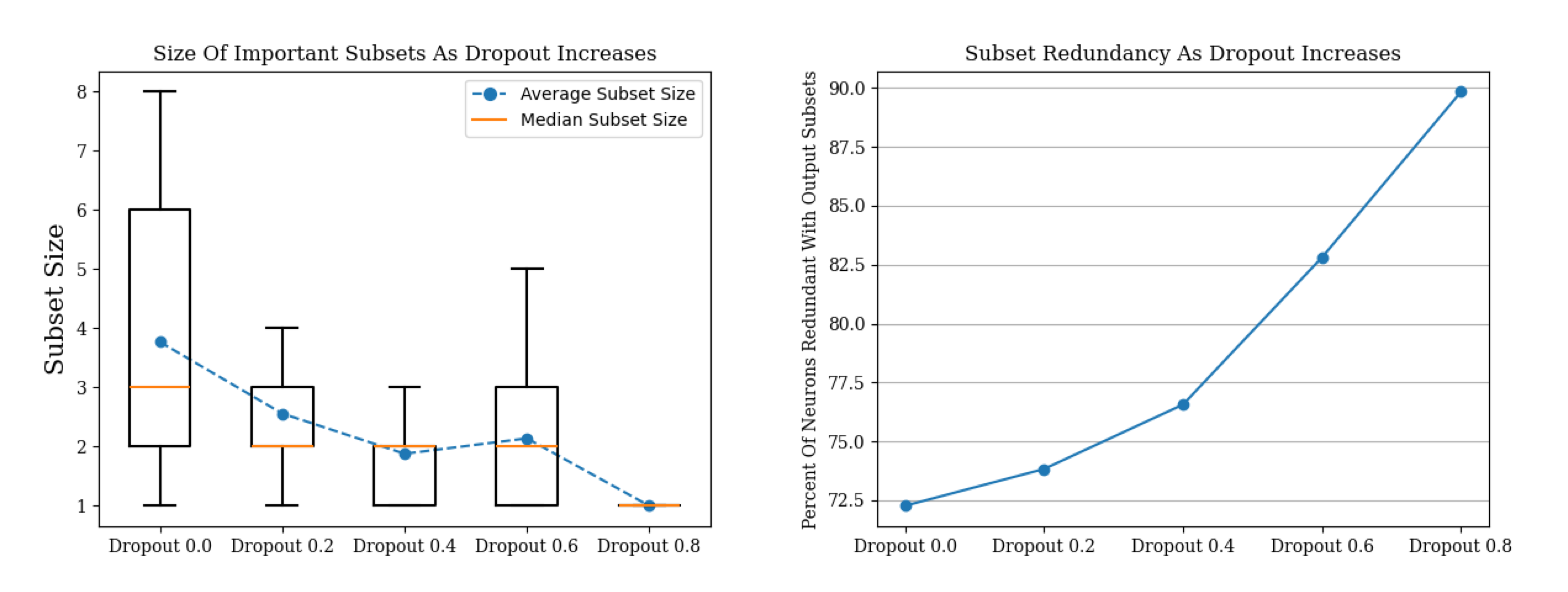}
    \caption{Left: Size of Subsets returned by ActSpec as a function of Dropout regularization. Right: Amount of Redundancy present as a function of Dropout regularization. }
    \label{fig:drop}
\end{figure*}

\section{EXPERIMENTS}

We begin with a number of synthetic tests where the exact representation of the function is known. This allows us to verify the behavior of our method and compare against existing approaches, highlighting strengths and weaknesses. An intuitive first step is to analyze a classical Boolean function. We introduce the function of interest and implement it using two neural networks. The first network has hard-coded weights and represents the function exactly, while the second network is trained until negligible reconstruction error is obtained. These different ways of encoding the function in a network affect the results of certain interpretability measures. After the initial set of comparisons, we modify the function by either adding constant variables or noise, both of which have no affect on the function output. The former case is introduced to showcase undesirable behavior from existing approaches, and the latter is meant to test accuracy and sample efficiency in higher-dimensional settings.

\begin{table}[h]
    \centering
    \begin{tabular}{c|c}
        $\pm \frac{1}{8}$ & $\emptyset, 1, 2, 3, 4, 13, 24, 123, 124, 134, 234, 1234$\\
        \midrule
         $\pm \frac{3}{8}$& $12, 14, 23$\\
         \midrule
         $\pm \frac{5}{8}$& $34$\\
    \end{tabular}
    \caption{Distribution of Fourier coefficients for the multi-tier function. Value of the coefficient on the left, associated collection of subsets with that value on the right.}
    \label{tab:tier}
\end{table}

After these synthetic tests, we move onto cases where the ground truth is unknown, starting with a simple MLP used to classify MNIST. Since MNIST is a dataset that can be naturally binarized, we begin by performing analysis on the input layer of the network. Doing so allows the results to be interpreted as pixel sets in the space of binarized images. We then move on to intermediate layers of the network, and showcase how ActSpec allows us to investigate and visualize the representational changes of Dropout regularization. Finally, we apply our tools to interpret the internals of a roBERTa model which was finetuned for sentiment analysis on the go-emotions dataset. 

\subsection{Multi-tiered Function}
The ground truth function for our first synthetic test is Boolean over four input variables, and defined such that $f(X_1, X_2, X_3, X_4) = 1$ if and only if either $X_3 = X_4 = 1$, $X_1 \geq X_2 \geq X_3 \geq X_4$, or if $X_1 \leq X_2 \leq X_3 \leq X_4$ \cite{tier}. It's Fourier spectrum is non-uniform, with higher concentration around the variables $X_3$ and $X_4$ in comparison to $X_1$ and $X_2$. Different values of the threshold parameter $\tau$ yield different collections of subsets, with the highest contribution coming from the joint state $X_3X_4$. For a more detailed presentation of the exact Fourier coefficients of this function, please see Table \ref{tab:tier} or the Appendix. In the absence of any noteworthy name for this function, we refer to it as a multi-tiered function, but the reason for choosing it over other such functions was arbitrary.

The Fourier coefficients returned by ActSpec are not immediately comparable to the outputs of most interpretability methods, making evaluation and comparison difficult. However, in the setting of pseudo-Boolean functions, a natural connection exists between Fourier coefficients and measures of variable importance. The natural measure of importance for pseudo-Boolean functions is Influence (average Sensitivity), which measures how often flipping the bit of $X_i$ leads to a change in the output $f(X)$ \cite{odonnell}. Surprisingly, the Influence of $X_i$ has an alternative form as the sum over all squared Fourier coefficients containing $X_i$. Thus, ActSpec can be used to estimate the Influence, which in turn can be compared with existing measures of importance and attribution. In order to do such a comparison, many approaches require access to the gradients and internals of some underlying network.

For our hard-coded multi-tier network, we explicitly compute each of the $16$ parities and take the appropriate weighted sum. This is accomplished with a hard-coded $6$ layer MLP with ReLU activations. The full architecture details are available in the Appendix. While this network is an exact representation of the underlying function, it's highly unlikely that such a network would be learned via gradient descent. Thus, our learned multi-tier network is a smaller $3$ layer MLP trained to near zero MSE $(2.24$e-$08)$. The ground truth importance for each variable $X_i$ is given by the Influence $Inf_i$, and deviation from this ground truth is measured using the total variation distance. Since most attribution methods are local \cite{loc-glob} each was averaged over all samples to match the averaged output of Influence. All attribution methods were sourced from the Captum library \cite{captum}. Table \ref{tab:synth} shows distance to the ground truth across a number of attribution approaches for both the hard-coded and learned network. In the hard-coded network, many methods systematically underestimated the importance of either $X_3$ or $X_4$. All methods performed as good or better on the learned network, with most approaches obtaining close to ground truth. ActSpec was able to perfectly estimate the Influence by outputting only the top 4 Fourier coefficients, since the Fourier density is otherwise uniform.

\subsubsection{Spurious Constant Variables}

To test a variation of the aforementioned synthetic experiment, we introduce a single spurious variable which is constant $(1)$ for each observed sample. This is an example of a function which is not observed over the full hypercube, and thus has a variable that is redundant with respect to the empty set (constant). Note that the mutual information between this variable and the function output is $0$. We use a learned network with $5$ input variables that is otherwise the same setup as the previous experiment. Across all attribution methods, the spurious variable was deemed important - often the most important - despite it providing no information about the output. Since ActSpec filters out constant and redundant variables, the spurious variable has no effect on its measure of importance. Such a result reveals a fundamental bias in attribution methods in favor of variables that are "active". 

\subsubsection{Spurious Noise Variables}

In the final synthetic experiment, we introduce $96$ spurious variables of random binary noise, making the total number of variables $100$. We test a high sample regime and a low sample regime, with $1000$ and $50$ samples respectively for each estimate. Under these constraints, ActSpec is not able achieve the ground truth exactly, but comes fairly close in the high sample case. When a low number of samples are used, not all important subsets are reliably discovered, with only the largest coefficients (such as $X_3X_4$) appearing consistently. Despite this, ActSpec is still able to perform better than all other methods, though Saliency comes fairly close. It should also be noted that Shapley Values and Shapley Sampling were not able to run in a reasonable amount of time due to the number of variables being too large. 
\begin{figure*}
    \centering
    \includegraphics[scale=0.2]{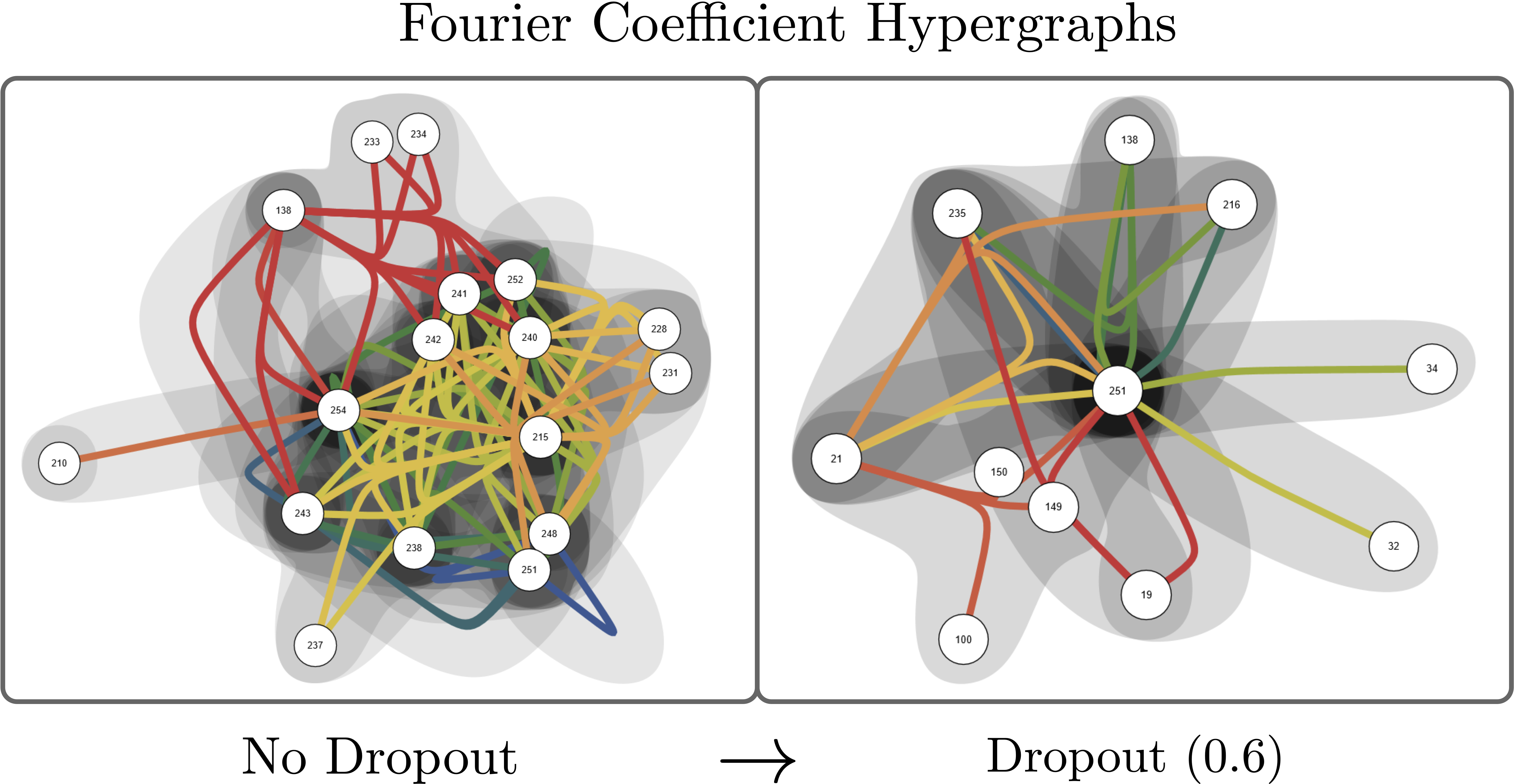}
    \caption{Hypergraph representations of sets output by ActSpec for the intermediate representations of an MNIST network with and without Dropout regularization}
    \label{fig:hyper}
\end{figure*}

\subsection{MNIST Input Layer}
With a set of baselines established for the performance of the algorithm in cases where the ground truth is known, we move onto the evaluation of neural networks. We begin with a standard 3-layer layer MLP with ReLU activations and Dropout trained to high accuracy. The pretrained weights used for the network can be found here: (anon, \cite{}). To start, ActSpec is run on the binarized MNIST input, with two separate types of MNIST dataset. The first dataset contains only two-digits, such as $1's$ and $7's$, or $0's$ and $8's$. The purpose of this is to evaluate which features ActSpec determines are important for distinguishing the two classes. The other experiments are performed on the full MNIST dataset, meaning the features returned are meaningful for identifying a digit from all others. Figure \ref{fig:mnist_input} visualizes the results of this experiment. If an input pixel is included in some output set of ActSpec, it is marked as red. If a variable is marked as redundant to any of theses sets, it is marked with orange. Together, the red and orange pixels represent the feature sets that contribute the most to the overall task. For distinguishing between a $0$ and an $8$ in the two-label setting, the most important features were those that make up the inner bar. Likewise, when distinguishing between a $7$ and $1$, the most important features are those associated with the top bar, and the width/ angle of the $7$. When moving on to the full dataset, the relevant features change significantly. For each digit, a collection of pixels that identified the general boundary were identified as important. Intuitively this makes sense, since in order to identify a digit from all others, you can no longer use features like the $7$'s top bar, since that also appears in other digits. We also plotted the average importance for Integrated Gradients and Saliency in the settings described. A red pixel in these plots means it is one of the top $100$ in terms of attribution for the method. Both attribution methods failed to account for the differences in the experimental setup, and returned roughly the same attributions in both cases.
\subsection{MNIST Intermediate Layer}
When moving from the input layer to an intermediate layer of a network, we lose the ability to give intuitive insight into what the features correspond to. To combat this, we performed a number of experiments with different levels of Dropout regularization, and show how information from the Fourier coefficients can be leveraged to understand how the underlying representation is changing. In Figure \ref{fig:drop} we show two intuitive results which connect Dropout and the Fourier representation. In the first experiment, we vary Dropout and measure the size of subsets returned by ActSpec. If a larger quantity of neurons are dropped during training, the network has to ensure that a smaller number of neurons contains the relevant information for prediction. Thus, we would expect subset size to shrink as the information concentration increases, which is what we observe. In the second experiment, we vary Dropout and measure the quantity of neurons that are redundant with respect to the output sets. As Dropout increases, the chance the relevant information is discarded also increases. As a preventative measure, it makes sense to represent the relevant information redundantly, which is corroborated by our results. Finally, Figure \ref{fig:hyper} shows a hypergraph visualization of the set of subsets returned by ActSpec in the case of no Dropout and relatively high Dropout. Each colored edge in the hypergraph corresponds to set membership for a collection of neurons. We see visually that the representation with Dropout is much less distributed than the representation without.

\subsection{Sentiment Analysis }
We run ActSpec on a pre-trained roBERTa model \cite{roberta} which was fine tuned for sentiment analysis on the go-emotions dataset \cite{go}. We chose this task because it has a natural classification output which is suited for our analysis. All estimates were computed with respect to the MLP in layer $10$ after the attention block. The intermediate layers of transformer-based networks take on a value for each token in the input sentence. This differs from the synthetic examples and the MNIST network, where each sample has a single intermediate output. To account for this, we perform our estimation in batches, where each batch is all tokens in a single sentence. Any change to an intermediate neuron is thus applied to every token in the sentence batch. Experimental details such as the number of samples and threshold value can be found in the Appendix. A concentrated collection of subsets was returned, with some of the sets having cardinality up to $4$. We wish to test if this information can be leveraged towards some actionable goal. In much of LLM interpretability\cite{bert}\cite{bert_geom} work, the importance of an internal component is established using interventions, such as in Causal mediation analysis \cite{edit1}\cite{edit2}
. In a similar vein, we use the output sets associated with high-valued Fourier coefficients as the starting point for ablation experiments. For each sample we check to see if an important neuron is activated. If it is, we set the output of that neuron to $0$. Otherwise, we set it to the mean activation of that neuron, as detailed in \cite{patch}. Out of the $43410$ sentences in the training set, we tested on a random sample of $6400$. Changes to at least one of the relevant subsets lead to a change in classification decision $32.4\%$ of the time. In $0.96\%$ of cases, an intervention at the level of individual neurons did not match the intervention at the group level. As a concrete example, for the sentence, "Klokslag 12 I find really enjoyable, horror movie podcast, Im not a horror fan but I find these dudes very entertaining", intervening on neuron $700$ in layer $10$ lead to a change in perceived sentiment, but intervening on the set $(436, 700, 744)$ and $(436, 700, 744, 761)$ did not. Since each intervention is applied independently, the changes to neurons $436$ and $744$ counteract the change to $700$. This is preliminary evidence that the joint state of these neurons contributes meaningfully to the output prediction.

\section{CONCLUSION}
In this work, we take the first steps towards applying pseudo-Boolean Fourier analysis to problems in neural network interpretability. This is accomplished by showing that established methods can be applied to In-distribution settings with modifications to the combinatorial procedure. We demonstrate that intuitive explanations can be produced experimentally in cases where other methods fail to do so. Further validation is required regarding the strengths, limitations, and scope of this approach, but we believe this to be a good starting point for further exploration.

\bibliography{bib_draft}{}
\bibliographystyle{plain}

\end{document}